\documentclass{article}

\usepackage{microtype}
\usepackage{graphicx}
\usepackage{booktabs} %

\usepackage{hyperref}

\usepackage[accepted]{icml2023}

\usepackage{amsmath}
\usepackage{amssymb}
\usepackage{mathtools}
\usepackage{amsthm}

\usepackage[capitalize,noabbrev]{cleveref}

\theoremstyle{plain}

\theoremstyle{definition}

\theoremstyle{remark}

\usepackage[textsize=tiny]{todonotes}

\usepackage{graphicx}
\usepackage{subcaption}
\usepackage{float}
\usepackage{enumitem}       %

\usepackage{amsmath}
\usepackage{amssymb}
\usepackage{mathtools}
\usepackage{amsthm}

\usepackage{booktabs}
\usepackage{longtable}
\usepackage{makecell}

\newcommand{\pstar}{p^{\star}}
\newcommand{\ptheta}{p_{\theta}}
\newcommand{\pthetat}[1]{p_{#1}}
\newcommand{\energy}{E_{\theta}}
\newcommand{\energyt}[1]{E_{#1}}
\newcommand{\KL}{\mathrm{KL}}
\newcommand{\flowpush}[2]{\lambda_{#1}^{#2}}
\newcommand{\R}{\mathbb{R}}
\def\rmd{\mathrm{d}}
\def\eqsp{\,}
\newcommand{\norm}[1]{\left\lVert#1\right\rVert}
\newcommand{\abs}[1]{\left\lvert#1\right\rvert}

\icmltitlerunning{Balanced Training of Energy-Based Models with Adaptive Flow Sampling}

\begin{document}

\twocolumn[
\icmltitle{Balanced Training of Energy-Based Models with Adaptive Flow Sampling}

\begin{icmlauthorlist}
\icmlauthor{Louis Grenioux}{x}
\icmlauthor{Éric Moulines}{x}
\icmlauthor{Marylou Gabrié}{x}
\end{icmlauthorlist}

\icmlaffiliation{x}{CMAP, CNRS, Ecole Polytechnique, Institut Polytechnique de Paris, 91120 Palaiseau, France}

\icmlcorrespondingauthor{Louis Grenioux}{louis.grenioux@polytechnique.edu}
\icmlcorrespondingauthor{Éric Moulines}{eric.moulines@polytechnique.edu}
\icmlcorrespondingauthor{Marylou Gabrié}{marylou.gabrie@polytechnique.edu}
\icmlkeywords{Energy-based models, Normalizing flows, MCMC, Sampling}

\vskip 0.3in
]

\printAffiliationsAndNotice{}  %

\begin{abstract}
    Energy-based models (EBMs) are versatile density estimation models that directly parameterize an unnormalized log density. Although very flexible, EBMs lack a specified normalization constant of the model, making the likelihood of the model computationally intractable. Several approximate samplers and variational inference techniques have been proposed to estimate the likelihood gradients for training. These techniques have shown promising results in generating samples, but little attention has been paid to the statistical accuracy of the estimated density, such as determining the relative importance of different classes in a dataset.
 In this work, we propose a new maximum likelihood training algorithm for EBMs that uses a different type of generative model, normalizing flows (NF), which have recently been proposed to facilitate sampling. Our method fits an NF to an EBM during training so that an NF-assisted sampling scheme provides an accurate gradient for the EBMs at all times, ultimately leading to a fast sampler for generating new data. 
\end{abstract}

\section{Introduction}
An Energy based model (EBM) defines a probability distribution over $x \in \mathcal{X} \subset \R^d$ by a parameterized \emph{energy function} $\energy: \R^d \to \R$ with parameters $\theta \in \Theta$ as
\begin{equation}\label{eq:def_ebm_classic}
 \ptheta(x) = \frac{1}{Z_{\theta}} \exp\left(-\energy(x)\right)\eqsp .
\end{equation}
These conceptually simple models are very flexible, since the functional class of $\energy$ is unrestricted, provided that it must ensure that $\exp(-\energy(x))$ is integrable. However, this flexibility comes at the price of an unknown normalization constant $Z_{\theta} = \int \exp(-E_{\theta}(x)) \rmd x$, which is difficult to calculate in practice, and the lack of a simple sampling procedure for the model. As a result, it is difficult to train EBMs by maximum likelihood methods and to use them as generative models.  %
To overcome these difficulties, various Monte Carlo Markov chains (MCMC) algorithms, from the simplest (e.g., \cite{hinton2012practical}) to the most complicated (e.g., \cite{bereux_learning_2023}), as well as various variational inference methods (VI) \cite{welling_new_2002,Gabrie2015, dai_exponential_2019,grathwohl2021no} were considered to approximate the likelihood gradients. Alternatively, weaker learning objectives have been proposed to avoid sampling $\ptheta(x)$, including Score Matching \cite{hyvarinen_estimation_2005,song_generative_2019}, Noise Constrative Estimation \cite{Gutmann2010}, and Minimum Stein Discrepancy \cite{Grathwohl2020} - see \cite{Song2021} for a recent methodological review of EBMs.

One of the main challenges common to all these training approaches is the correct estimation of the density of multimodal distributions, that is, datasets with several different clusters in the data. This is due to the difficulty for MCMCs and VIs to accurately represent multimodal distributions, or the difficulty for weaker learning objectives that rely on scores $\nabla_x \log \ptheta(x)$ to capture this information (see, e.g., \cite{song_generative_2019}, Section 3.2.1). Meanwhile, the highly flexible parameterization of EBMs makes them particularly well suited to the multimodal setting, compared to the competing class of generative models parameterized as push-forward distributions, such as generative adversarial networks (GANs) or normalizing flows (NFs), whose ability to partition mass into multiple modes is inherently limited \cite{cornish_relaxing_2020,salmona_can_2022}.

However, a number of recent works have shown that some generative push-forward models, namely NFs, facilitate the sampling of multimodal distributions (see references in \cref{sec:background}).
Building on these results, we propose joint learning of an EBM with a companion NF, which allows efficient sampling of the EBM at any point in the training and, as a result, accurate maximum likelihood training of the EBM. As described in the Related Work section, several proposals have already been made to combine EBMs with NFs. Our works goes a step further in this direction by employing a calibrated NF-assisted MCMC \cite{Gabrie2022,Samsonov2022} recently shown to be particularly robust in the multimodal setting \cite{Grenioux2023}.

\section{Background}
\label{sec:background}
\paragraph{EBM maximum likelihood training} 

Given a training data distribution $\pstar$, the EBM log-likelihood can be written as $\ell_{\rm EBM}(\theta) = \mathbb{E}_{\pstar}[\log\ptheta(X)]$. This quantity is intractable due to the unknown $Z_\theta$ of \cref{eq:def_ebm_classic}, which translates into an expectation over $\ptheta$ in its gradient:
\begin{equation}\label{eq:grad_ebm}
    \nabla_{\theta} \ell_{\rm EBM}(\theta) = \mathbb{E}_{\pstar}[\nabla_{\theta} \energy(X)] - \mathbb{E}_{\ptheta}[\nabla_{\theta} \energy(X)].
\end{equation}
A Monte-Carlo estimation of $\nabla_{\theta} \ell_{\rm EBM}(\theta)$ requires training samples $x^{(+)}_i \sim \pstar(x)$, commonly referred to as \emph{positive samples} in the EBM context, and samples from the current model $x^{(-)}_i \sim \ptheta(x)$, respectively called \emph{negative samples}. Collecting $n$ samples of these two kinds yields the approximation for gradient \eqref{eq:grad_ebm} $\widehat{\nabla_{\theta} \ell_{\rm EBM}}(\theta, \{x^{(-)}_i, x^{(+)}_i\}_{i=1}^{n}) =$ 
\begin{align}\label{eq:update_theta}
     - \frac{1}{n} \left( \sum_{i=1}^n \nabla_{\theta} \energyt{\theta_k}(x^{(-)}_i) -  \sum_{i=1}^n \nabla_{\theta} \energyt{\theta_k}(x^{(+)}_i)\right).
\end{align}
Yet, obtaining exact samples from $p_{\theta_k}$ requires converging an MCMC, which is a costly procedure to repeat.
As a result, approximate sampling procedures have been proposed: in contrastive divergence (CD) \cite{Hinton2002}, a fixed small number of MCMC steps is ran starting from training samples at each gradient computation.
In \emph{persistent} CD (PCD), this simple idea was further refined by propagating the MCMC chains the negative samples across gradient updates
\cite{tieleman_training_2008}. 
For real valued-valued EBMs, CD and PCD most commonly employ \emph{Uncalibrated Langevin Algorithm} (ULA) \cite{Roberts1996}, a
local gradient-based sampler, which at step $k$ updates $x^{(k)}$ as
\begin{equation} \label{eq:langevin}
    x^{(t+1)} = x^{(t)} - \eta \nabla \log \energy(x^{(t)}) + \sqrt{2\eta} z^{(t)}\eqsp
\end{equation}
where $\eta$ is the step size of the algorithm and $z^{(t)} \sim \mathcal{N}(0,I)$. 

If ULA 
samples the target distribution $\ptheta$ asymptotically in time, it typically cannot converge in a manageable number of iterations for distribution that are multimodal. 
While recent research suggests that using a non-convergent MCMC for drawing negative samples does not compromise sample quality if a consistent sampling scheme is employed during and after training \cite{Nijkamp2019,Nijkamp2020, Nijkamp_Hill_Han_Zhu_Wu_2020, An2021, xie2022a}, 
it is not guaranteed that an EBM trained in this fashion captures the overall mass distribution between different modes (see the motivating example of \ref{sec:motivating_example}).

\paragraph{NF-Assisted sampling}\label{sec:nf_within_mcmc}

\emph{Normalizing flows} (NF) combine a \emph{base} distribution $\rho$ on $\R^d$ and a bijective transport map $T_{\alpha} : \R^d \to \R^d$ with parameters  $\alpha \in \mathbb{A}$ to define a generative model with density:
\begin{align}
    \flowpush{T_\alpha}{\rho}(x) = \rho(T_\alpha^{-1}(x)) \abs{J_{T_\alpha^{-1}}(x)},
\end{align}  
from which samples are straightforwardly obtained as $X = T_{\alpha}(Z)$ with $Z \sim \rho$. NFs can be trained on training data to maximize the explicit likelihood. 
We point the reader to the reviews \cite{Papamakarios2021,Kobyzev2021}. 

Thanks to their tractable densities and direct sampling procedure, NFs have found applications in statistical inference either as a variational family \cite{rezende_variational_2015,wu_solving_2019} or as helpers in sampling algorithms \cite{parno_transport_2018,albergo_flow-based_2019,noe_boltzmann_2019, muller_neural_2019,mcnaughton_boosting_2020,hackett_flow-based_2021} (among others). 
Given a target distribution $\pi$, known up to a normalization constant, the general idea of NF-assisted inference is to train the map $T_\alpha$ such that $\flowpush{T_\alpha}{\rho}$ approaches $\pi$. In this context, since no training sample is available a priori, the flow is trained either by minimizing the reverse Kulleback-Leibler (KL) divergence $\KL(\flowpush{T_{\alpha}}{\rho} || \pi) = \mathbb{E}_{\flowpush{T_\alpha}{\rho}}[ \log \flowpush{T_\alpha}{\rho}(x) / \pi(x) ] $ \cite{rezende_variational_2015}
or through an adaptive MCMC procedure maximizing a proxy of the likelihood \cite{parno_transport_2018,mcnaughton_boosting_2020, naesseth_markovian_2021, Gabrie2022}. When $\pi$ is multimodal, the reverse KL is prone to mode collapse \cite{jerfel_variational_2021,hackett_flow-based_2021}, while the adaptive MCMC training can leverage prior knowledge of the different modes' basins to seed the learning of a model covering all the regions of interest. 
Once trained, samples from the flow $\flowpush{T_\alpha}{\rho}$ approximating the target $\pi$ can be debiased via importance sampling or various MCMC schemes. A recent comparative study \cite{Grenioux2023} shows that methods based on independent proposals from the flow, such as independent Metropolis-Hastings (e.g., \cite{nicoli_asymptotically_2020}), are more robust to sample multimodal distributions compared to re-parametrization schemes such as neutra-MCMC \cite{Hoffman2019neutra}. In the context of EBM training, these observations suggest the use a of flow-based adaptive MCMC using independent proposals, FlowMC (\cref{algo:flowmc} in \cref{app:algorithms}), to provide accurate negative samples.

\section{EBMs with Flow Sampling}\label{sec:coupled_ebm}

\begin{figure*}
    \centering
    \begin{subfigure}{0.49\linewidth}
        \centering
        \includegraphics[width=\linewidth]{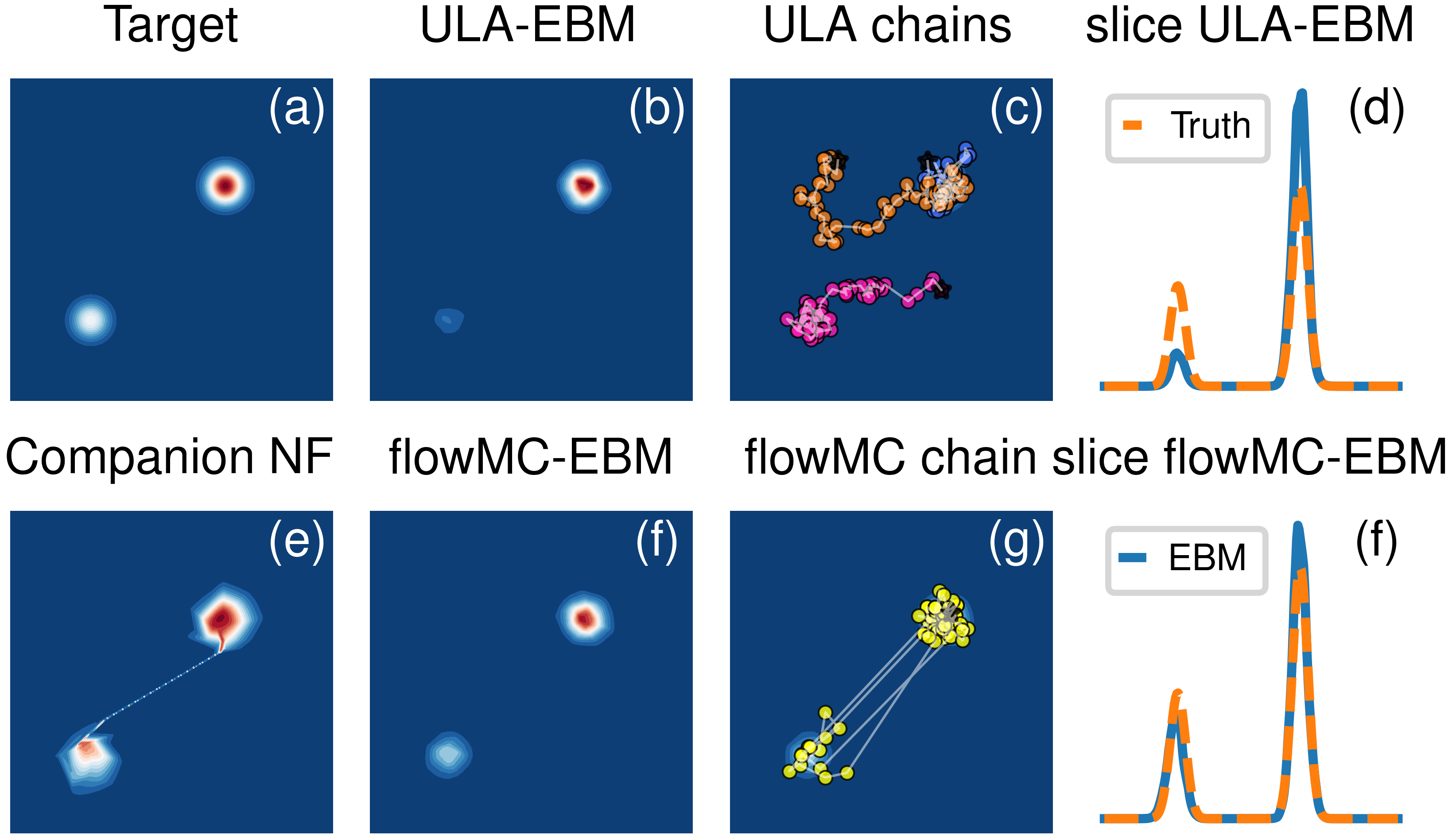}
    \end{subfigure}%
    \begin{subfigure}{0.49\linewidth}
        \includegraphics[width=\linewidth]{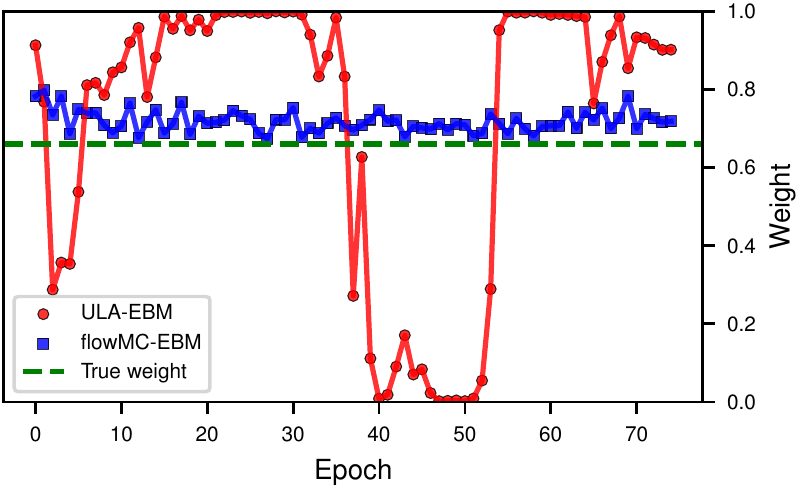}
    \end{subfigure}
    \caption{\textbf{Comparison of PCD and flowMC EBM training on toy 2d mixture with 2:1 weight ratio.} The EBM learned with flowMC \textbf{(f)} captured correctly the relative weights, unlike the EBM trained with ULA \textbf{(b)}, as is also clear from the conditional densities along the axis going trough the centroids (\textbf{d,h}). This statistical accuracy is promoted by the fast mixing NF-assisted sampling \textbf{(g)}. \textbf{(Right) Estimated weight of the top right mode during the training of ULA-EBM and flowMC-EBM} 
    (Details in \cref{app:motivating_example})}
    \label{fig:exp:motivating_example:tutorial_and_weight}
    \vspace{-0.3cm}
\end{figure*}

We suggest to train a NF to maintain a good overlap between the flow's $\flowpush{T_{\alpha}}{\rho}$ and EBM's $\ptheta$ throughout training so as to implement an NF-assisted sampler to draw negative samples (\cref{algo:ebm_coupled}).
For this symbiosis to work in practice, that is $\flowpush{T_{\alpha_t}}{\rho} \approx \pthetat{\theta_t}^{\rho}$ at all times, we 
slightly modify the EBM definition \cref{eq:def_ebm_classic} using the flow base distribution $\rho$ as a tilt\footnote{Note that this change does not modify the gradient of $\ell_{\rm EBM}(\theta)$ (see \cref{eq:grad_ebm}) see proof in \cref{app:proofs}.}: 
\vspace{-0.2cm}
\begin{equation}\label{eq:def_ebm_new}
    \ptheta^{\rho}(x) = \frac{1}{Z_{\theta}} \exp\left(-\energy(x)\right) \rho(x) \eqsp.
\end{equation}
Choosing initially $\theta_0$ such that $\energyt{\theta_0}(\cdot) = 0$ and $\alpha_0$ such that $T_{\alpha_0}(\cdot) = {\rm Id}(\cdot)$ leads to a perfect equality at initialization $\pthetat{\theta_0}^{\rho} = \rho = \flowpush{T_{\alpha_0}}{\rho}$. Then, the learning rates $\gamma_{\rm EBM}$ and $\gamma_{\rm flow}$ in \cref{algo:ebm_coupled} need to be co-adjusted for the matching to be approximately maintained all along.

Unlike strategies using ULA to obtain negative samples, our proposition is statistically reliable as it uses a calibrated MCMC sampler which handles multimodality. Thanks to the good agreement between $\flowpush{T_{\alpha_t}}{\rho}$ and $\pthetat{\theta_t}^{\rho}$, non-local moves proposed by the NF in the flowMC sampler allow rapid-mixing between modes. This coupled learning of two generative models takes the best of both: the constrained but tractable NF approximates an unconstrained but intractable EBM. Additionally, it provides a natural and efficient way to sample the resulting EBM through FlowMC.

\begin{algorithm}[tb]
    \caption{flowMC-EBM training step on data distribution $\rho^{\star}$ with persistent initialization}
    \label{algo:ebm_coupled}
    \begin{algorithmic}
    \STATE {\bfseries Input:} 
    EBM parameter $\theta_k$, flow parameters $\alpha_k$, learning rates $\gamma_{\rm EBM}$ and $\gamma_{\rm flow}$, batch size $n$, local step size $\eta$, number of MCMC steps $N$, persistent state $\{\tilde{x}_i\}_{i=1}^n$
    \STATE {\bfseries Output:} $\theta_{k+1}$, $\alpha_{k+1}$, updated $\{\tilde{x}_i\}_{i=1}^n$ 

    \STATE \underline{1. Draw positive samples} $\{x^{(+)}_i\}_{i=1}^n \sim \rho^{\star}$
    \STATE  \underline{2. Draw negative samples from $\pthetat{\theta_k}$ 
    using flow $T_{\alpha_k}$}
    $$
        x^{(-)}_i = \mathrm{flowMC}(\pthetat{\theta_k}^{\rho}, \tilde{x}_i, T_{\alpha_k}, \rho, \eta, N) \text{ for all } i
    $$
    \STATE \underline{3. Update the persistent state} $\{\tilde{x}_i\}_{i=1}^n = \{x^{(-)}_i\}_{i=1}^n$
    \STATE  \underline{4. Perform EBM gradient descent step} 
    $$
    \theta_{k +1} = \theta_k - \gamma_{\rm EBM} \widehat{\nabla_{\theta} \ell_{\rm EBM}}(\theta_k, \{x^{(-)}_i, x^{(+)}_i\}_{i=1}^{n})
    $$
    \STATE \underline{5. Perform EBM gradient ascent on NF likelihood}
    $$ %
        \alpha_{k+1} = \alpha_k + \gamma_{\rm flow} \left(\frac{1}{n} \sum_{i=1}^n \nabla_{\alpha} \ln \flowpush{T_{\alpha_k}}{\rho}(x^{(-)}_i)\right)
    $$\;
    \vspace{-0.5cm}
    \end{algorithmic}
\end{algorithm}

\paragraph{Related works}

Several directions have already been explored combining energy based models with push-forward generative models to leverage their complementary strenghts.   
\cite{Xie_Lu_Gao_Wu_2018, Xie_Zheng_Li_2021} suggested to use a Variational Autoencoder (VAE) and \cite{xie2022a} a NF (CoopFlow algorithm) to provide initial negative samples before running short chains during maximum likelihood training, 
reporting good sample quality but no guarantee of statistical accuracy.

Closer to this work in their concern to resort to a calibrated and converged sampler (NT-EBM algorithm) \cite{xiao2020exponential, nijkamp2022mcmc} leverage an alternative type of NF-assisted sampler using the flow bijective mapping as a preconditionner \cite{parno_transport_2018,Hoffman2019neutra}. 
Yet recent work suggest that a multimodal problem remains so when re-parametrized by a flow and therefore that chains mixing is not guaranteed \cite{Grenioux2023}. 
This approach was also performed using VAE in \cite{xiao2021vaebm}. Another occurrence of a tilted EBM with a pushforward model was also explored with Generative Adversarial Networks \cite{arbel2021generalized} using a Langevin sampler in latent space  without assessment of the statistical performance on multimodal datasets.

Lastly, a set of works considered the simulatenous learning of an auxiliary model for sampling along with the EBM using the Fenchel dual description of the intractable partition function \cite{dai_exponential_2019,grathwohl2021no}.Yet this strategy amounts to minimizing the reverse KL objective, prone to mode-collapse, and the statistical robustness of the methods to multimodal targets was untested.

\section{Numerical experiments}

\paragraph{Motivating example}\label{sec:motivating_example}

We first illustrate the difficulty of learning relative weights with persistent ULA-EBM training on a 2D mixture of Gaussians (\cref{fig:exp:motivating_example:tutorial_and_weight}). 
The incapacity of ULA to mix between the modes ((c) versus (g)), introduces a biais in the estimation of the gradient \eqref{eq:grad_ebm}, which leads to an over-correction of mismatched weights: ULA-EBM entirely entirely erases a mode multiple times during training, before recreating it and the final weight at which the EBM training stops is not a robust estimation of the target weights (\cref{fig:exp:motivating_example:tutorial_and_weight} Right).
In flowMC-EBM training on the other hand, calibrated negative samples lead to a stable estimation of the weights during learning and a final accurate density estimation ((f) and (h)).
The companion flow (e), more constrained in its parametrization, does not achieve a fit as accurate as the EBM, yet its match with the EBM remains good enough to facilitate the fast-mixing MCMC key to success.   
A detailed comparison including more algorithms from Related Works is also presented in \cref{app:motivating_example_advanced}.

\begin{figure}[t]
    \centering
    \begin{subfigure}[b]{\linewidth}
        \centering
        \includegraphics[width=\linewidth]{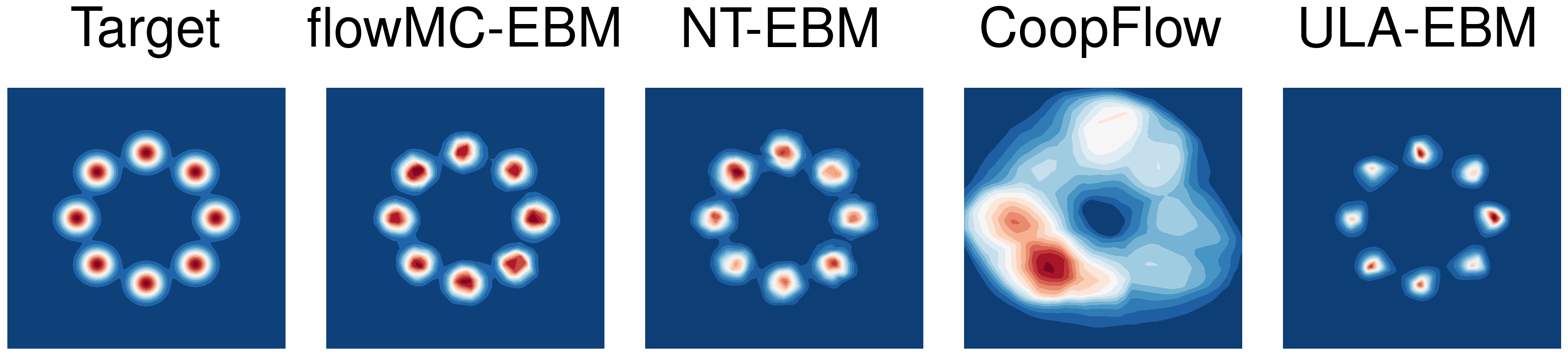}
    \end{subfigure}
    \begin{subfigure}[b]{\linewidth}
        \centering
        \includegraphics[width=\linewidth]{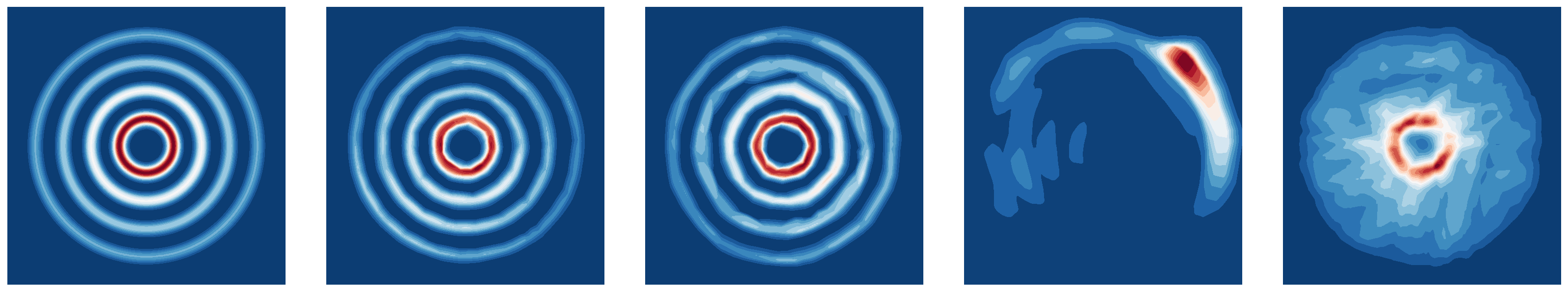}
    \end{subfigure}
    \caption{Estimated energies using different algorithms on toy 2D experiments - \textbf{(Top)} 8 Gaussians \textbf{(Bottom) } Rings}
    \label{fig:exp:2d_toys:densities}
\end{figure}

\begin{table}[t]
    \caption{Median squared error on log-density
    ($\underset{x}{\operatorname{med}}(\log\rho_{\theta}(x) - \log\rho^{\star}(x))^2)$. Best metrics in bold. \vspace{-0.5cm}}
    \label{table:exp:2d_toys:mse}
    \begin{center}
        \begin{small}
            \begin{sc}
                \begin{tabular}{lll}
                \toprule
                 & 8 Gaussians & Rings \\
                \midrule
                ULA-EBM & 1.86 & 5.39 \\
                NT-EBM & 0.97 & 0.62 \\
                CoopFlow & 58.75 & 7.78 \\
                flowMC-EBM & \textbf{0.94} & \textbf{0.40} \\
                \bottomrule
                \end{tabular}
            \end{sc}
        \end{small}
    \end{center}
    \vspace{-0.5cm}
\end{table}
\paragraph{2D distributions with more modes and complicated geometries.} \label{sec:toy_2d}We benchmark approaches combining EBMs and NFs on the 2D distributions \emph{8-Gaussians} and \emph{rings}. The different models shared the same EBM/flow architecture and were trained for the same number of iterations. The final densities displayed in \cref{fig:exp:2d_toys:densities} highlight that our algorithm outperforms competitors in weighting the different modes. This is quantitatively confirmed by the energy errors computed in \cref{table:exp:2d_toys:mse}. See \cref{app:toy_2d} for more details.
\paragraph{High dimensional mixture} \label{sec:four_gaussians}

\begin{table}[t]
    \caption{Maximum $\hat{R}$ across dimension of negative samples on Gaussian mixture computed on 128 independent chains started from the persistent state (or from the flow for CoopFlow).  \cref{fig:app:four_gaussians:rhat} in \cref{app:four_gaussians} ) }
    \label{table:exp:four_gaussians:rhat}
    \begin{center}
        \begin{small}
            \begin{sc}
            \begin{tabular}{llll}
                \toprule
                 & Dim. 16 & Dim. 32 & Dim. 64 \\
                \midrule
                CoopFlow & 20.52 & 44.12 & 51.66 \\
                NT-EBM & 2.02 & 2.50 & 3.10 \\
                ULA-EBM & 7.30 & 9.29 & 90.90 \\
                flowMC-EBM & \textbf{1.01} & \textbf{1.01} & \textbf{1.05} \\
                \bottomrule
            \end{tabular}
            \end{sc}
        \end{small}
    \end{center}
    \vspace{-0.5cm}
\end{table}

We now consider an equally-weighted mixture of 4 Gaussians in dimensions 16, 32 and 64. We compare here again flowMC-EBM with NT-EBM and CoopFlow, yet focusing this time on characterizing the mixing of the chains throughout learning. Using identical EBM/flow architectures trained for the same number of iterations,
we report the $\hat{R}$ metric of the negative chains for each model at the end of training in \cref{table:exp:four_gaussians:rhat} for the end of training and in \cref{app:four_gaussians} throughout training. Meant to compare the intra-chain variance and the inter-chains variance, reaching a $\hat{R}$ close to 1 is a necessary criteria of convergence of an MCMC \cite{Vehtari2021}. flowMC-EBM is the only algorithm allowing proper mixing. See \cref{app:four_gaussians} for more details.

\paragraph{CIFAR 10} Given our computational budget, we were able to train a flowMC-EBM producing samples of medium quality (see \cref{fig:app:cifar10:samples} of \cref{app:cifar10} along with training details). Nonetheless, negative chains mix between modes as the companion NF proposal's are accepted around 5-10\% of the time. Given the number of parameters reported in related work, we expect that a more expressive flow and energy parametrization would improve the outcome.  

\section{Conclusion}

By combining an EBM and NF, we manage to tackle the generative models trilemma described in \cite{xiao2022tackling}. The trilemma states that among the desirable properties of (i) fast sampling, (ii) high-quality samples and (iii) mode-coverage/diversity of the produced samples, a generative model typically only features two out of three. Our numerical experiments show that the cost of training two models is compensated by obtaining a strategy without compromises with respect to the three aspects. Going even further than mode-coverage, we show that our algorithm enables a precise evaluation of the mode relative weights, a topic rarely discussed in the literature.   

\section*{Acknowledgements}

M.G. thanks Eric Vanden-Eijnden for insightful discussions. L.G. and M.G. acknowledge funding from Hi! Paris. The work was partly supported by ANR-19-CHIA-0002-01 “SCAI”. Part of this research has been carried out under the auspice of the Lagrange Center for Mathematics and Computing.

\bibliography{refs_marylou,refs_louis}
\bibliographystyle{icml2023}

\newpage
\appendix
\onecolumn

\section{Algorithms}
\label{app:algorithms}

We recall useful algorithms: the Unagjusted Langevin (ULA) in \cref{algo:ula}, the persistent ULA-EBM training in \cref{algo:ebm_classic}, the Metropolis Adjusted Langevin Algorithm (MALA) in \cref{algo:mala} and finally the flowMC sampler in \cref{algo:flowmc}. We also recall the Iterated Sampling Importance Resampling (i-SIR) in \cref{algo:isir} which can be used as a drop-in replacement of steps \emph{1.a}-\emph{1.b} of the flowMC algorithm (\cref{algo:flowmc}) as suggested by \cite{Samsonov2022}.

\begin{algorithm*}[tb]
    \caption{Unadjusted Langevin algorithm (ULA) with target distribution $\pi$}
    \label{algo:ula}
    \begin{algorithmic}
        \STATE {\bfseries Input:} $x^{(0)}$ initial sample, $\eta$ step-size, $L$ number of MCMC steps
        \STATE {\bfseries Output:} $(x^{(k)})_{k=1}^L$ samples according to $\pi$
        \WHILE{$k < L; k = 1$}
            \STATE $x^{(k)} \sim \mathcal{N}(x^{(k-1)} + \eta \nabla \log \pi (x^{(k-1)}), 2 \eta I_d)$
        \ENDWHILE
    \end{algorithmic}
\end{algorithm*}

\begin{algorithm*}[tb]
    \caption{ULA-EBM training step with persistent initialization  \cite{du_implicit_2019,Nijkamp_Hill_Han_Zhu_Wu_2020} }
    \label{algo:ebm_classic}
    \begin{algorithmic}
        \STATE {\bfseries Input:} Data distribution $\pstar$, EBM parameter $\theta_k$, learning rate $\gamma$, batch size $n$, ULA step size $\eta$, number of ULA steps $L$, persistent state $\{\tilde{x}_i\}_{i=1}^n$
        \STATE {\bfseries Output:} New EBM parameter $\theta_{k+1}$
        \STATE \underline{1. Draw positive samples} $\{x^{(+)}_i\}_{i=1}^n \sim \pstar$
        \STATE \underline{2. Draw negative samples}
            $$
            x^{(-)}_i = \mathrm{ULA}(-E_{\theta_k}, \tilde{x}_i, \eta, L) \text{ for } i = 1 \dots n \;
            $$ 
        \STATE \underline{3. Update persistent state} 
            $$
            \{\tilde{x}_i\}_{i=1}^n = \{x^{(-)}_i\}_{i=1}^n 
            $$
        \STATE \underline{4. Perform a GD step on $-\ell_{\rm EBM}(\theta)$ (cf \cref{eq:grad_ebm})\;}
    \end{algorithmic}
\end{algorithm*}

\begin{algorithm*}[tb]
    \caption{Metropolis-adjusted Langevin algorithm (MALA) with target distribution $\pi$}
    \label{algo:mala}
    \begin{algorithmic}
        \STATE {\bfseries Input:} $x^{(0)}$ initial sample, $\eta$ step-size, $L$ number of MCMC steps
        \STATE {\bfseries Output:} $(x^{(k)})_{k=1}^L$ samples according to $\pi$
        \WHILE{$k < L; k = 1$}
            \STATE \underline{1. Sample the local proposal} $x^{(k)} \sim \mathcal{N}(x^{(k-1)} + \eta \nabla \log \pi (x^{(k-1)}), 2 \eta I_d)$
            \STATE \underline{2. Metropolis-Hastings accept-reject}
            $x^{(k)} = x^{(k-1)} \text{ with prob. }$ $ 1-\min\left[1, \frac{\pi(x^{(k)}) q(x^{(k-1)} | x^{(k)})}{\pi(x^{(k-1)}) q(x^{(k)} | x^{(k-1)})}\right]$\\
            where $q(x'|x) = \exp(-\norm{x' - x - \eta \nabla \log \pi(x)}^2 / (4\eta))$
        \ENDWHILE
    \end{algorithmic}
\end{algorithm*}

\begin{algorithm*}[tb]
    \caption{FlowMC adaptive sampling of target distribution $\pi$ \cite{Gabrie2022}}
    \label{algo:flowmc}
    \begin{algorithmic}
        \STATE {\bfseries Input:} $x^{(0)}$ initial sample, $T_{\alpha}$ initial flow, $\rho$ base distribution, $\eta$ MALA step-size, $\gamma$ flow learning rate, $n_{\rm MALA}$ number of MALA steps, $L$ number of MCMC steps
        \STATE {\bfseries Output:} $(x^{(k)})_{k=1}^L$ samples according to $\pi$
        \WHILE{$k < N; k = 1$}
            \STATE \underline{1.a Sample the flow} $x^{(k)} \sim \flowpush{T_\alpha}{\rho}(x)$
            \STATE \underline{1.b Metropolis-Hastings accept-reject}
            $x^{(k)} = x^{(k-1)} \text{ with prob. }$ $ 1-\min\left[1, \frac{\pi(x^{(k)})\flowpush{T_\alpha}{\rho}(x^{(k-1)})}{\pi(x^{(k-1)})\flowpush{T_\alpha}{\rho}(x^{(k)})}\right]$
            \STATE \underline{2. Sample with MALA from $x^{(k)}$}
            $x^{(k+1:k+n_{\rm MALA}+1)} = \mathrm{MALA}(\log \pi, x^{(k)}, \eta, n_{\rm MALA})$\;
            $k = k + 1 + n_{\rm MALA}$\;
            \STATE \underline{3. Likelihood ascent step on the flow}
            $\alpha = \alpha + \gamma \sum_{k'<k} \ln \flowpush{T_\alpha}{\rho}(x^{(k')}) $
        \ENDWHILE
    \end{algorithmic}
\end{algorithm*}

\begin{algorithm*}[tb]
    \caption{Iterated Sampling Importance Resampling (i-SIR) \cite{Rubin1987}}
    \label{algo:isir}
    \begin{algorithmic}
        \STATE {\bfseries Input:} $x^{(0)}$ initial sample, $\lambda$ proposal distribution, $N$ number of particles, $L$ number of MCMC steps
        \STATE {\bfseries Output:} $(x^{(k)})_{k=1}^L$ samples according to $\pi$
        \WHILE{$k < L; k = 1$}
            \STATE \underline{1. Draw a pool of proposals} $y_{2:N}^{(k+1)} \sim \lambda$
            \STATE \underline{2. Set the $x^{(k)}$ as the first element of the pool} $y_1^{(k+1} = x^{(k)}$ 
            \STATE \underline{3. Compute the importance weights}
            $w_i^{(k+1)} = w(y_i^{(k+1)}) / \sum_{j=1}^N w(y_j^{(k+1)})$ for all $i = 1, \ldots, N$ where $w = \pi / \lambda$
            \STATE \underline{4. Select the next state}
            $i^{(k+1)} \sim \mathcal{M}(w_1^{(k+1)}, \ldots, w_N^{(k+1)})$
            \STATE \underline{5. Update the next state}
            $x^{(k+1)} = y_{i^{(k+1)}}^{(k+1)}$
        \ENDWHILE
    \end{algorithmic}
\end{algorithm*}

\section{Gradients of tilted distribution} \label{app:proofs}

This section provides the proof of the remark at the beginning of \cref{sec:coupled_ebm}. Let's consider a tilted EBM as in \cref{eq:def_ebm_new} with $\energyt{\theta} : \R^d \to \R$ a parametrized energy function, $Z_{\theta}$ the associated normalizing constant of $\ptheta^{\rho}$ and $\rho$ the base distribution. We have that $\nabla_{\theta} \log \ptheta^{\rho}(x) = -\nabla_{\theta} \energyt{\theta}(x) - \nabla_{\theta} \log Z_{\theta}$. The second term can be expressed as an expectation over $\ptheta^{\rho}$
\begin{align*}
    \nabla_{\theta} \log Z_{\theta} &= \nabla_{\theta} \log \int \exp(-\energyt{\theta}(x)) \rho(x) \rmd x \\
    &= \left( \int \exp(-\energyt{\theta}(x)) \rho(x) \rmd x \right)^{-1} \nabla_{\theta} \int \exp(-\energyt{\theta}(x)) \rho(x) \rmd x \\
    &= \left( \int \exp(-\energyt{\theta}(x)) \rho(x) \rmd x \right)^{-1} \int \nabla_{\theta} \exp(-\energyt{\theta}(x)) \rho(x) \rmd x \\
    &= \left( \int \exp(-\energyt{\theta}(x)) \rho(x) \rmd x \right)^{-1} \int \exp(-\energyt{\theta}(x)) (-\nabla_{\theta} \energyt{\theta}(x)) \rho(x) \rmd x \\
    &= \int \left( \int \exp(-\energyt{\theta}(y)) \rho(y) \rmd y \right)^{-1} \exp(-\energyt{\theta}(x)) (-\nabla_{\theta} \energyt{\theta}(x)) \rho(x) \rmd x \\
    &= \int (-\nabla_{\theta} \energyt{\theta}(x)) \ptheta^{\rho}(x) \rmd x \\
    &= \mathbb{E}_{\ptheta^{\rho}}[-\nabla_{\theta} \energyt{\theta}(X)]\eqsp.
\end{align*}
The gradient of the maximum likelihood objective can then be formulated as 
\begin{align*}
    \nabla_{\theta} \ell_{\rm EBM}(\theta) &= \mathbb{E}_{\pstar}[\nabla_{\theta}\log\ptheta(X)]\\
    &=  \mathbb{E}_{\pstar}[-\nabla_{\theta} \energyt{\theta}(X) - \nabla_{\theta} \log Z_{\theta}] \\
    &= -\mathbb{E}_{\pstar}[\nabla_{\theta} \energyt{\theta}(X)] + \mathbb{E}_{\ptheta^{\rho}}[\nabla_{\theta} \energyt{\theta}(X)]\eqsp.
\end{align*}

\section{Experiments}

\subsection{Common remarks on implementation of the different EBM training algorithms throughout the numerics}

While Hamiltonian Monte Carlo (HMC) can be used  instead of Langevin as a local MCMC sampler \cite{nijkamp2022mcmc}, we stick to MALA in our implementations of CoopFlow, NT-EBM and flowMC-EBM for two reasons : (i) it is much faster than HMC \footnote{Only one gradient per generated sample.} (ii) it makes the comparison between models easier. Moreover, most of the EBM training algorithms in the literature originally used only a few MCMC steps at each iteration. In our benchmark of these algorithms, we typically use more MCMC steps.
Finally, we used specific version of CoopFlow and ULA-EBM algorithms. 

\paragraph{Coop Flow}
For CoopFlow \cite{xie2022a} we used the "best" version of their algorithm where the flow is pre-trained on samples from $\pstar$. During the coupled training of the EBM, the pre-trained flow will be initially frozen for less than 5\% of the total training time. We also disabled the noise in Langevin steps as advised by the authors. 

\paragraph{ULA-EBM} For the ULA-EBM, the hyper-parameters (custom temperature and learning rate) were selected using the recommendations from \cite{Nijkamp_Hill_Han_Zhu_Wu_2020}.

\paragraph{FlowMC with i-SIR updates} In practice, we use the flowMC algorithm (\cref{algo:flowmc}) with the i-SIR algorithm (\cref{algo:isir}) as a global sampler (i.e. for steps \emph{1.a} to \emph{1.b} of the flowMC algorithm). As it accepts one flow proposal from a large pool of samples, this algorithm enables shorter decorelation times provided enough memory and parallel computation capabilities.

\subsection{Motivating example}\label{app:motivating_example}

The first example from \cref{sec:motivating_example} uses an unequilibrated mixture of two multivariate normal distributions in 2D
$$
    \pstar(x) = \frac{1}{3} \mathcal{N}(-1.5\mathbf{1}_2, 0.05 I_2) + \frac{2}{3} \mathcal{N}(+1.5\mathbf{1}_2, 0.1 I_2)\eqsp
$$
where $\mathbf{1}_d$ is the vector with only unit coordinates in dimension $d$. Both models were trained for 75 epochs on a dataset of length 16384 using a persistent state of size 1024 and a batch size of 64 with a learning rate of 0.01 (same one for the flow and the EBM). The global steps are i-SIR steps with 32 particles. The Langevin samplers used a step size of 0.01. The classic EBM used 10 MCMC steps at each step and the flowMC-EBM used the same number of steps decomposed with 2 global steps between two sequence of 4 local steps. We used the same parameterization for the EBM as in \cite{Nijkamp_Hill_Han_Zhu_Wu_2020} (about 10k parameters). The flow is a RealNVP \cite{Dinh2017} with 4 layers each one featuring MLPs with 16 neurons on 3 layers (about 1k parameters total).

\subsection{Extended motivating example}\label{app:motivating_example_advanced}

This additional motivating example aims at building intuition on why other methods are failing at relative weights estimation. This experiment uses an 2D unequilibrated mixture of four Gaussians $\pstar = \sum_{i=1}^4 w_i \mathcal{N}(\mu_i, 0.05 I_2)$ where the $\mu_i$ are evenly distributed on a horizontal axis between $x = -3$ and $x = 3$ and $w = (0.1, 0.2, 0.3, 0.4)$.

\begin{table}[t]
    \begin{minipage}[b]{0.45\linewidth}
        \centering
        \captionof{figure}{Estimated energies using different algorithms on the extended motivating example.}
        \label{fig:app:motivating_example_extended:ebms}
        \includegraphics[width=\linewidth]{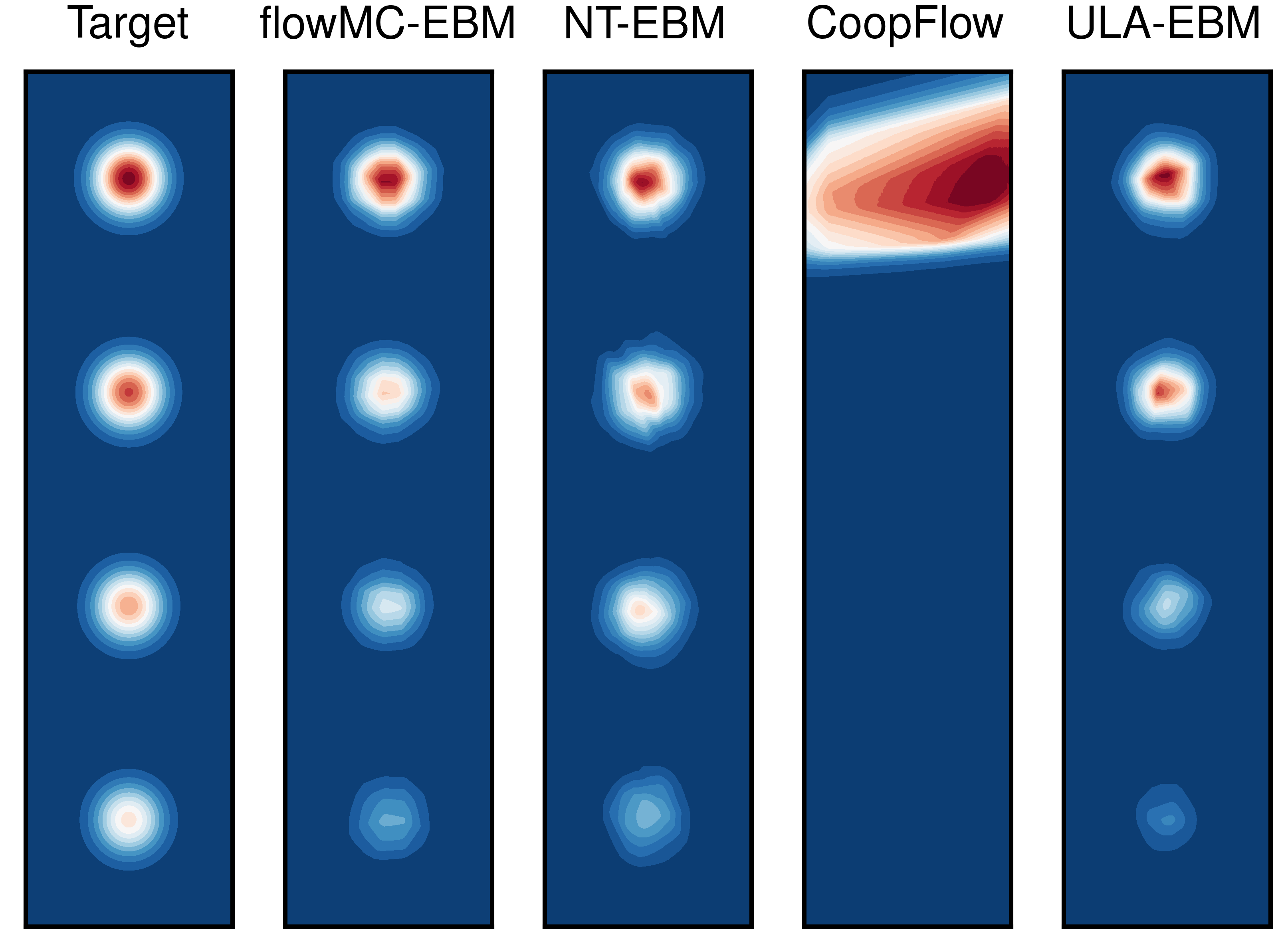}
    \end{minipage}
    \hspace{1em}
    \begin{minipage}[b]{0.45\linewidth}
        \centering
        \captionof{figure}{Space partitioning for histograms - the level lines corresponds to $\pstar$ and the four colors correspond to each zone.}
        \label{fig:app:motivating_example_extended:zones}
        \includegraphics[width=\linewidth]{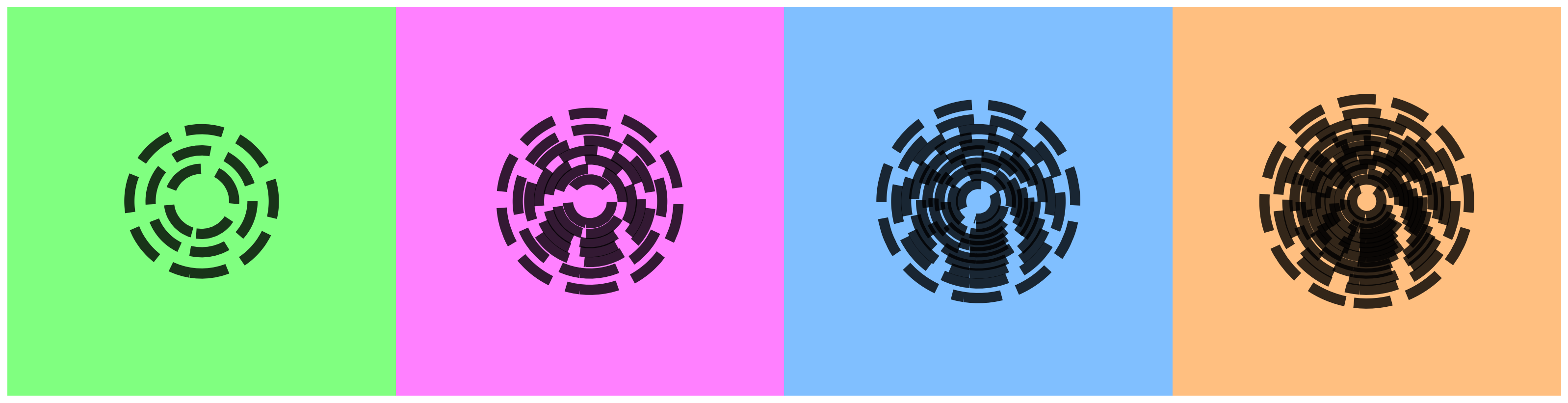}
        \centering
        \captionof{table}{Mean Squared Errors (MSE) after training different EBMs 
 algorithms on the extended motivating example.}
        \label{table:app:motivating_example_extended:metrics}
        \begin{small}
            \begin{tabular}{c | c | c | c }
                \toprule
                \thead{Algorithm} & \thead{MSE of the \\init. wrt $\energyt{\theta}$} & \thead{MSE of the neg. \\ samples wrt $\energyt{\theta}$} & \thead{MSE of $\energyt{\theta}$\\wrt $-\log\pstar$} \\
                \midrule
                ULA-EBM & 1.19e-03 & 1.79e-03 & 1.24e-03 \\
                CoopFlow & 1.21e-01 & 1.11e-01 & 1.24e-01 \\
                NT-EBM & 9.27e-04 & 3.85e-04 & 9.87e-04 \\
                \textbf{flowMC-EBM} & \textbf{5.19e-04} & \textbf{3.12e-05} & \textbf{4.29e-04} \\
                \bottomrule
            \end{tabular}
        \end{small}
    \end{minipage}
\end{table}

The obtained EBMs are presented in \cref{fig:app:motivating_example_extended:ebms}. \cref{table:app:motivating_example_extended:metrics} provides a quantitative comparison between the different algorithms. The mean squared error of samples $X$ with respect to distribution $p$ are computed by partitioning the space in as much regions as the number of modes of $p$ (see \cref{fig:app:motivating_example_extended:zones}) and then building an histogram of the samples $X$ depending on which zone each sample is. The error is computed against the weights of the distribution $p$. For the distribution $\energyt{\theta}$, the weights are computed by manually summing the mass in each region corresponding to the modes of $\pstar$.

The second column of \cref{table:app:motivating_example_extended:metrics} represents the quality of the initialisation of the negative chains with respect to the current EBM - this is the error that CoopFlow should minimize by using a flow. The third column represents the error of the negative samples with respect to the current EBM - this is the error that NT-EBM should minimize by mixing in the latent space. \cref{table:app:motivating_example_extended:metrics} shows that our algorithm is the only one minimizing both errors leading to a better EBM (see the fourth column). Our algorithm improves the initialization by using calibrated MCMC algorithms and improves negative samples by using the companion normalizing flow to visit even more modes during the sampling phase.

\begin{figure}[t]
    \centering
    \includegraphics[width=0.75\linewidth]{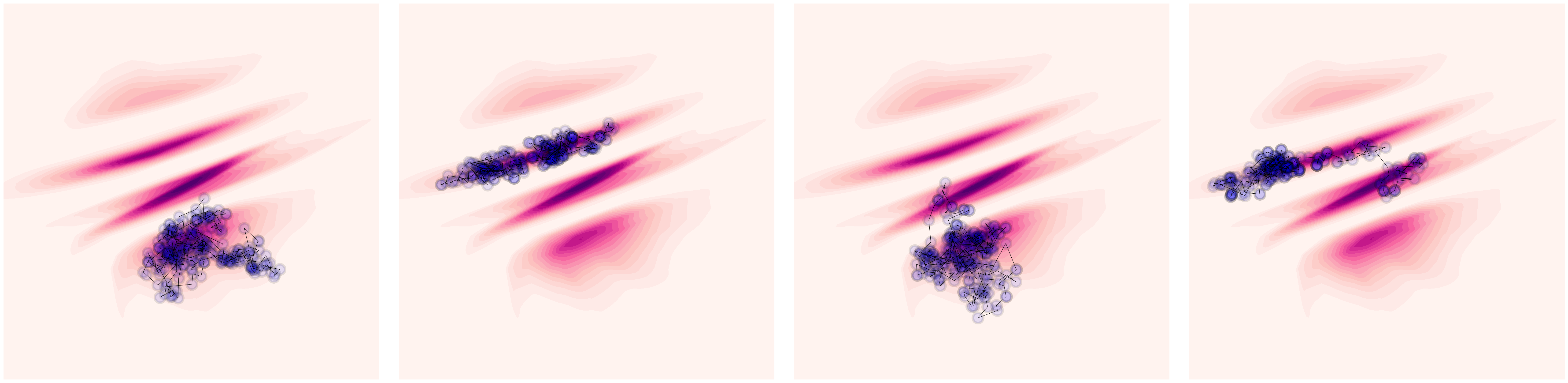}
    \caption{Four different negative chains viewed in the latent space of NT-EBM trained on the extended motivating example}
    \label{fig:app:motivating_example_extended:latent_samples}
\end{figure}

The failure of CoopFlow at providing good initialization is likely due to the fact that pre-training the flow on data provide bad starting samples if the current EBM's weights are not close to the data weights : at the end of training, the weights of its flow are $(0.11, 0.20, 0.29, 0.40)$ which is very accurate with respect to data but the MSE with respect to the EBM is of order $10^{-1}$. This is worsen by the lack of any MCMC calibration on the initialisation and the negative samples. The failure of NT-EBM at improving its initialization error is likely due to the fact that the local sampler isn't mixing in the latent space (as suggested by \cite{Grenioux2023}) as shown in \cref{fig:app:motivating_example_extended:latent_samples}.

The hyper-parameters of each algorithm are summarized in \cref{table:app:toy_2d:ebm_params}. The \textit{Use only last step ?} column correspond to whether the negative samples are the last elements of the MCMC chains or not. The learning rates is $10^{-2}$ for EBM and NT-EBM and $10^{-3}$ for CoopFlow and our algorithm (flow and EBM). The dataset is 60000 samples large with a batch size of 256 and a persistent size of 8192. The training lasted 100 epochs for each algorithm. The EBM and the flow are the same is in \cref{app:motivating_example}. When required, the flows were pretrained for 1024 steps at a $10^{-2}$ learning rate with the reverse KL objective. The global sampler is i-SIR with 64 particles.

\begin{table}[t]
    \caption{EBM hyper-parameters for the extended motivating example}
    \label{table:app:motivating_example_extended:ebm_params}
    \begin{center}
        \begin{small}
            \begin{tabular}{c | c | c | c | c }
                \toprule
                \thead{Algorithm} & \thead{MCMC Sampler} & \thead{\# MCMC steps} & \thead{Use only last step ?} & \thead{Starting strategy}\\
                \midrule
                ULA-EBM & \texttt{ula} with $(1.25 \times 10^{-1})^2 / 2$ step size & $512$ & yes & uniform \\
                NT-EBM & \texttt{mala} with target acceptance $75\%$ & $128$ & yes & flow \\
                CoopFlow & \texttt{ula} with $5 \times 10^{-4}$ step size & $256$ & yes & flow \\
                flowMC-EBM & \texttt{flowMC} with 1 global step and 24 \texttt{mala} local steps & $128$ & no & flow \\
                \bottomrule
            \end{tabular}
        \end{small}
    \end{center}
\end{table}

\subsection{2D distributions with more modes and complicated geometries.}\label{app:toy_2d}

The 8 Gaussians distribution is a mixture of 8 equally weighted Gaussians $\mathcal{N}(\mu_i, 0.15^2 I_2)$ where $\mu_i = (\cos(2\pi t_i), \sin(2\pi t_i))$ and $t_i = i/8$ with $i \in \{0, \ldots, 7\}$. The rings distributions is the inverse polar reparametrization of a distribution $p_z$ which has itself a decomposition into two univariate marginals $p_r$ and $p_{\theta}$. $p_r$ is a mixture of 4 Gaussians $\mathcal{N}(i+1, 0.15^2)$ with $i \in \{0, \ldots, 3\}$ describing the radial position and $p_{\theta}$ is a uniform distribution over $[0, 2\pi]$ which describe the angular position of the samples.

The hyper-parameters of each algorithm are summarized in \cref{table:app:toy_2d:ebm_params}. The \textit{Use only last step ?} column correspond to whether the negative samples are the last elements of the MCMC chain or not. If \textit{subsampling by k} is mentioned in the this column, it means that we divided the number of parallel MCMC chains by $k$ and subsampled the resulting chains evenly $k$ times. The learning rates was $10^{-2}$ for EBM and NT-EBM and $10^{-3}$ for CoopFlow and our algorithm (flow and EBM). The dataset was 60000 samples large with a batch size of 256 and a persistent size of 8192. The training lasted 100 epochs for each algorithm. The EBM and the flow are the same is in \cref{app:motivating_example}. When required, the flows were pretrained for 1024 steps at a $10^{-2}$ learning rate with the reverse KL objective. The global sampler is i-SIR with 64 particles.

\begin{table}[t]
    \caption{EBM hyper-parameters for 8 Gaussians / Rings experiments}
    \label{table:app:toy_2d:ebm_params}
    \begin{center}
        \begin{small}
            \begin{tabular}{c | c | c | c | c }
                \toprule
                \thead{Algorithm} & \thead{MCMC Sampler} & \thead{\# MCMC steps} & \thead{Use only last step ?} & \thead{Starting strategy}\\
                \midrule
                ULA-EBM & \texttt{ula} with $(1.25 \times 10^{-1})^2 / 2$ step size & $512$ & yes & uniform \\
                NT-EBM & \texttt{mala} with target acceptance $75\%$ & $128$ & yes & flow \\
                CoopFlow & \texttt{ula} with $5 \times 10^{-4}$ step size & $256$ & yes & flow \\
                flowMC-EBM & \texttt{flowMC} with 1 global step and 127 \texttt{mala} local steps & $128$ & subsampling by 4 & flow \\
                \bottomrule
            \end{tabular}
        \end{small}
    \end{center}
\end{table}

\subsection{High-dimensional example}\label{app:four_gaussians}

\begin{figure}[t]
    \centering
    \includegraphics{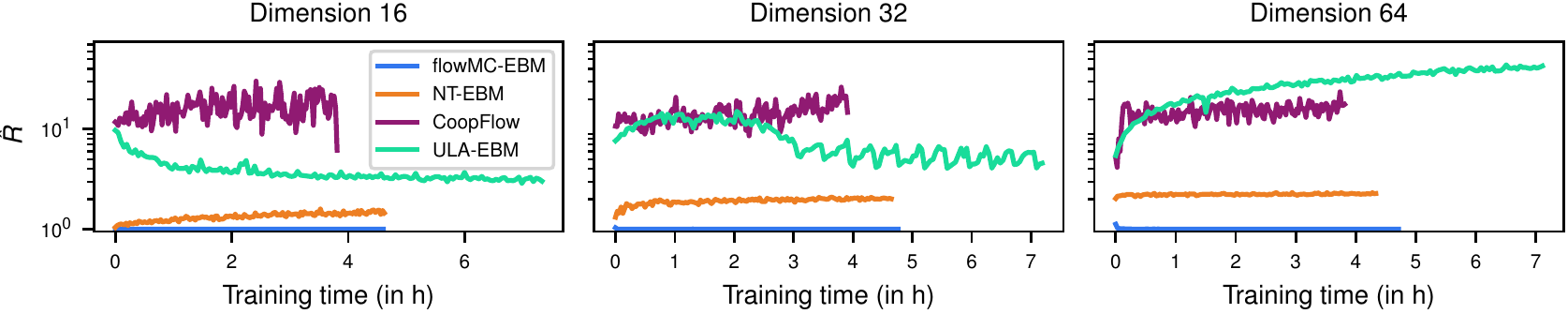}
    \caption{Extended version of figure \cref{table:exp:four_gaussians:rhat} where the $\hat{R}$ is displaying along training time}
    \label{fig:app:four_gaussians:rhat}
\end{figure}

In this experiment, $\pstar$ is a mixture of 4 equally weighted isotropic Gaussians $\pstar_i = \mathcal{N}(\mu_i, I_d)$ where the $\mu_i$ are defined as $\mu_1 = a \times (1, 1, 1, \ldots, 1, 1, 1)$, $\mu_2 = a \times (-1, -1, -1, \ldots, 1, 1, 1)$, $\mu_2 = - \mu_2$ and $\mu_4 = -\mu_1$ and $a = 0.5919$. This specific value of $a$ guarantees that if $X \sim \pstar_i$ then $\forall j \neq i, \mathbb{P}(\|X - \mu_j\| < \|X - \mu_i\|) \leq 10^{-10}$ for any dimension $d$.

The hyper-parameters of each algorithm are summarized in \cref{table:app:four_gaussians:ebm_params}. The \textit{Use only last step ?} column correspond to whether the negative samples are the last elements of the MCMC chains or not. If \textit{stacked} is mentionned in the \textit{\# MCMC steps} column, it means that the $k_g$ global steps and $k_l$ local steps are considered as a single MCMC step. The learning rates was $10^{-2}$ for EBM and NT-EBM and $10^{-3}$ for CoopFlow and our algorithm (flow and EBM). The dataset was 50000 samples large with a batch size of 128 and a persistent size of 8192. The training lasted 150 epochs for each algorithm. The EBM parametrization is the same as the one described in \cite{Nijkamp_Hill_Han_Zhu_Wu_2020}. The global sampler is i-SIR with 128 particles. The flows used here are RealNVPs. The base of the flow is $\rho = \mathcal{N}(0, \sigma^2 I_d)$ where $\sigma^2$ is the maximum variance of $\pi$ along each dimension. All the coupling layers have 3 hidden layers initialized with very small weights $(\simeq 10^{-6})$. The other hyper-parameters can be found in table \cref{table:app:four_gaussians:flows}. When required, the flows were pretrained for 512 steps at a $10^{-2}$ learning rate with the reverse KL objective.
 
\begin{table}[t]
    \caption{RealNVP architecture for the high-dimensional Gaussian mixture}
    \label{table:app:four_gaussians:flows}
    \begin{center}
        \begin{small}
            \begin{tabular}{c | c | c | c | c | c}
                \toprule
                \thead{Dimension} & \thead{Size of hidden layers} & \thead{\# RealNVP blocks} \\
                \midrule
                16 & 32 & 2 \\
                32 & 39 & 2 \\
                64 & 51 & 2 \\
                \bottomrule
            \end{tabular}
        \end{small}
    \end{center}
\end{table}

\begin{table}[t]
    \caption{EBM hyper-parameters for the high-dimensional mixture experiment}
    \label{table:app:four_gaussians:ebm_params}
    \begin{center}
        \begin{small}
            \begin{tabular}{c | c | c | c | c }
                \toprule
                \thead{Algorithm} & \thead{MCMC Sampler} & \thead{\# MCMC steps} & \thead{Use only last step ?} & \thead{Starting strategy}\\
                \midrule
                ULA-EBM & \texttt{ula} with $10^{-3}$ step size & $512$ & yes & gaussian \\
                NT-EBM & \texttt{mala} with target acceptance $75\%$ & $128$ & yes & flow \\
                CoopFlow & \texttt{ula} with $10^{-4}$ step size & $256$ & yes & flow \\
                flowMC-EBM & \texttt{flowMC} with 1 global step and 8 \texttt{mala} local steps & $16$ (stacked) & no & flow \\
                \bottomrule
            \end{tabular}
        \end{small}
    \end{center}
\end{table}

\subsection{Image distribution}\label{app:cifar10}

\begin{figure}[t]
    \centering
    \includegraphics{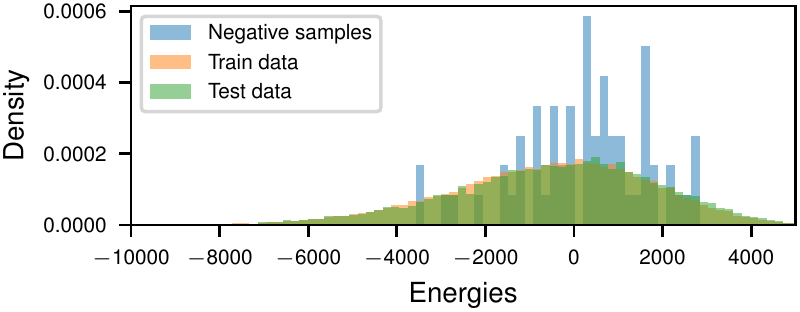}
    \caption{Energy histograms of our algorithm trained on CIFAR10}
    \label{fig:app:cifar10:hists}
\end{figure}

\begin{figure}[t]
    \centering
    \includegraphics[width=\linewidth]{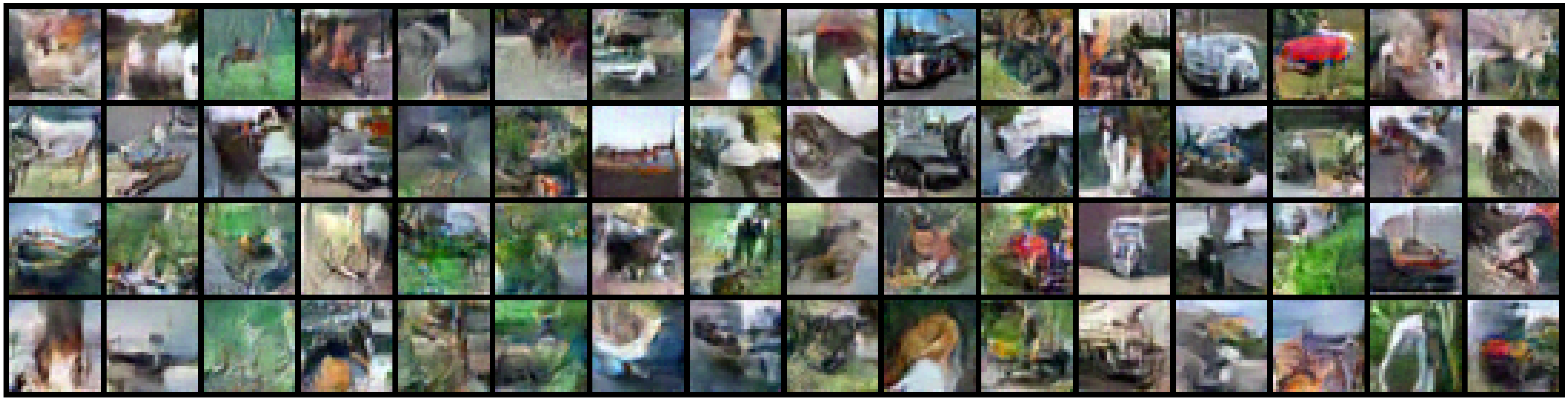}
    \caption{Samples from our algorithm trained on the CIFAR10 dataset - We display the last state of 64 MCMC chains of 512 steps sampled in parallel. The chains were initialized in the persistent state.}
    \label{fig:app:cifar10:samples}
\end{figure}

We tested our algorithm on the CIFAR10 dataset (see training details below). Given our computational budget, the resulting generative model can produce samples of medium quality (see \cref{fig:app:cifar10:samples}). The global sampler proposals are getting accepted 5-10\% of the time. Given number of parameters reported in related work, we expect that a more expressive flow and energy parametrization would improve the outcome. Following the experiments of \cite{dai_exponential_2019}, we compute the energy distribution of the generated samples and compare it against the data samples (see \cref{fig:app:cifar10:hists}). The EBM produces sample in the right energy range but not as diverse as the training data. 

The flowMC-EBM was trained on the CIFAR10 dataset using a batch size of 128 and for 70 epochs and a persistent size of 1024. The learning rates of the flow and the EBM was $10^{-4}$ with a decay by $0.99$ every 50 batches. We used an L2 regularization with coefficient $10^{-4}$ on the EBM loss i.e., we added a $\ell_{\text{reg}}$ term where $\ell_{\text{reg}}(\theta, \{x^{(-)}_i, x^{(+)}_i\}_{i=1}^n) = \lambda \sum_{i=1}^n (\energyt{\theta}(x^{(-)}_i))^2 + (\energyt{\theta}(x^{(+)}_i))^2$ with $\lambda = 10^{-4}$. The sampler stacked 1 i-SIR global step with 128 particles and 32 local steps of MALA three times in total to build the negative samples.

The dataset was centered between -1 and 1 and also dequantized (ie. if $x$ is an image with integer pixels in range $[0,255]$ then its dequantization is $\tilde{x} = (255x + U) / 256$ where $U \sim \mathcal{U}([0,1])$). The dataset was augmented with random horizontal flipping.

The EBM used was the one described in \cite{du_implicit_2019} and used in \cite{xie2022a} with a smaller last MLP layer (about 4.5M parameters only). The flow is a RealNVP taken from \href{https://github.com/fmu2/realNVP}{this Github repository} with \texttt{base\_dim} to 64 and 6 residual blocks (about 4M parameters). The flow was stacked with an affine transformation bringing the images in $[0,1]$ and the logisitic transformation described in \cite{Dinh2017}. The base of the flow is a standard centered multivariate Gaussian $\mathbf{N}(0,I_{3\times32\times32})$ and the prior of the EBM (see \cref{eq:def_ebm_new}) was pushed through the additional transformations stacked onto the flow.

\end{document}